\documentclass[letterpaper, 10 pt, conference]{ieeeconf}

\IEEEoverridecommandlockouts

\usepackage{amsmath,amsfonts}
\usepackage{algorithmic}
\usepackage{algorithm}
\usepackage{array}
\usepackage[caption=false,font=normalsize,labelfont=sf,textfont=sf]{subfig}
\usepackage{textcomp}
\usepackage{stfloats}
\usepackage{url}
\usepackage{verbatim}
\usepackage{graphicx}
\usepackage{cite}
\usepackage{xcolor}
\usepackage{hyperref}

\hypersetup{
    colorlinks=true,       
    linkcolor=red,         
    citecolor=green,       
    filecolor=red,         
    urlcolor=black         
}

\usepackage{multicol}
\usepackage{footnote} 
\usepackage{tabularx}
\usepackage{booktabs}
\usepackage{color}
\usepackage{bigstrut}
\usepackage{bbding}
\usepackage{amssymb}
\usepackage{pifont}
\usepackage{multirow}
\usepackage{cases}

\usepackage{bbding}
\usepackage{utfsym}
\usepackage{thmtools}

\definecolor{lightgray}{rgb}{0.83, 0.83, 0.83}
\definecolor{forestgreen}{rgb}{0.13, 0.55, 0.13}

\usepackage{colortbl}
  
\usepackage[utf8]{inputenc} 
\usepackage[T1]{fontenc}    

\definecolor{lightgray}{rgb}{0.83, 0.83, 0.83}
\definecolor{forestgreen}{rgb}{0.13, 0.55, 0.13}

\newcommand{\mypara}[1]{\vspace{1mm}\noindent\textbf{#1}}
\newcommand{\name}{NS-MAE}

\newcommand{\nus}{nuScenes}

\definecolor{codegreen}{rgb}{0,0.6,0}
\definecolor{codegray}{rgb}{0.5,0.5,0.5}
\definecolor{codepurple}{rgb}{0.58,0,0.82}
\definecolor{backcolour}{rgb}{0.95,0.95,0.92}

\definecolor{MyGreen}{RGB}{0, 180, 0}
\definecolor{MyRed}{RGB}{180, 0, 0}
\definecolor{MyBlue}{RGB}{30, 0, 180}

\definecolor{place3d_blue}{RGB}{66, 133, 244}
\definecolor{place3d_red}{RGB}{231, 66, 52}
\definecolor{place3d_yellow}{RGB}{251, 189, 5}
\definecolor{place3d_green}{RGB}{51, 168, 82}
\definecolor{place3d_gray}{RGB}{165, 165, 165}

\definecolor{building}{RGB}{70, 130, 180}
\definecolor{fence}{RGB}{0, 0, 230}
\definecolor{other}{RGB}{255, 255, 153}
\definecolor{pedestrian}{RGB}{255, 102, 178}
\definecolor{pole}{RGB}{255, 153, 153}
\definecolor{road_line}{RGB}{224, 224, 224}
\definecolor{road}{RGB}{224, 0, 224}
\definecolor{sidewalk}{RGB}{0, 204, 204}
\definecolor{vegetation}{RGB}{0, 204, 102}
\definecolor{vehicle}{RGB}{102, 102, 255}
\definecolor{wall}{RGB}{0, 128, 255}
\definecolor{traffic_sign}{RGB}{255, 51, 51}
\definecolor{ground}{RGB}{255, 204, 255}
\definecolor{bridge}{RGB}{255, 127, 80}
\definecolor{rail_track}{RGB}{96, 96, 96}
\definecolor{guard_rail}{RGB}{96, 0, 96}
\definecolor{traffic_light}{RGB}{255, 0, 0}
\definecolor{static}{RGB}{0, 153, 153}
\definecolor{dynamic}{RGB}{255, 255, 102}
\definecolor{water}{RGB}{0, 128, 255}
\definecolor{terrain}{RGB}{102, 102, 0}

\let\labelindent\relax
\usepackage{enumitem}

\title{\LARGE \bf 
Learning Shared RGB-D Fields: Unified Self-supervised Pre-training for Label-efficient LiDAR-Camera 3D Perception}

\author{Xiaohao Xu$^{1}$ \ Ye Li$^{1}$ \ Tianyi Zhang$^{2}$ \ Jinrong Yang$^{1}$ \ Matthew Johnson-Roberson$^{2}$ \ Xiaonan Huang$^{1}$
\thanks{$^{1}$Xiaohao Xu, Ye Li, Jinrong Yang, and Xiaonan Huang are with the Robotics Department, University of Michigan, Ann Arbor, MI 48109, USA. {\tt\small xiaohaox@umich.edu; xiaonanh@umich.edu}.}
\thanks{$^{2}$Tianyi Zhang and Matthew Johnson-Roberson are with the Robotics Institute, Carnegie Mellon University, Pittsburgh, PA 15213, USA.}
}
\begin{document}

\maketitle
\thispagestyle{empty}
\pagestyle{plain}

\begin{abstract}
Constructing large-scale labeled datasets for multi-modal perception model training in autonomous driving presents significant challenges. This has motivated the development of self-supervised pretraining strategies. However, existing pretraining methods mainly employ distinct approaches for each modality. In contrast, we focus on LiDAR-Camera 3D perception models and introduce a unified pretraining strategy, NeRF-Supervised Masked Auto Encoder (NS-MAE), which optimizes all modalities through a shared formulation. 
NS-MAE leverages NeRF’s ability to encode both appearance and geometry, enabling efficient masked reconstruction of multi-modal data.
Specifically, embeddings are extracted from corrupted LiDAR point clouds and images, conditioned on view directions and locations. Then, these embeddings are rendered into multi-modal feature maps from two crucial viewpoints for 3D driving perception: perspective and bird's-eye views. The original uncorrupted data serve as reconstruction targets for self-supervised learning.
Extensive experiments demonstrate the superior transferability of NS-MAE across various 3D perception tasks under different fine-tuning settings. Notably, NS-MAE outperforms prior SOTA pre-training methods that employ separate strategies for each modality in BEV map segmentation under the label-efficient fine-tuning setting. Our code is publicly available at \url{https://github.com/Xiaohao-Xu/Unified-Pretrain-AD/}.

\end{abstract}

\section{Introduction}
Multi-modal perception models~\cite{waymo2020,choudhary2024talk2bev,liu2022bevfusion} enhance the robustness of autonomous driving by integrating data from various sensors, such as LiDAR and multi-view cameras~\cite{li2024influence}. However, training these models in a conventional fully-supervised paradigm poses significant challenges, particularly in collecting large-scale datasets with accurately aligned 3D labels for both LiDAR point clouds and multi-view images. 
Progress in self-supervised learning~\cite{he2022masked} has alleviated the data demands of complex network architectures, enabling more scalable and data-efficient training~\cite{Vaswani2017AttentionIA,lin2024bev}.

\begin{figure}[t!]
\centering
\includegraphics[width=0.48\textwidth]{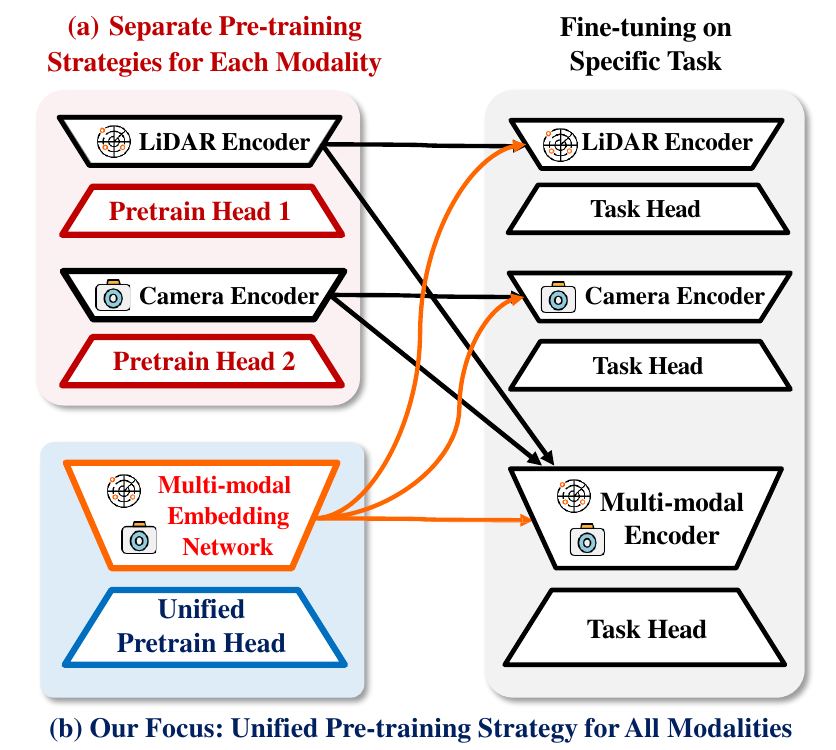} 
\caption{\textbf{Motivation.} (\textbf{a}) Existing methods predominantly focus on single-modal pre-training~\cite{voxel-mae,boulch2023also,yang2023gd}, or use distinct strategies for each modality in multi-modal models~\cite{bevdistill,li2022simipu,calico}. (\textbf{b}) Our unified self-supervised pre-training framework consistently applies a single approach across all modalities, improving the transferability of multi-modal representations.}
\label{fig:motivation}
\vspace{-4mm}
\end{figure}

In autonomous driving, self-supervised strategies~\cite{voxel-mae,boulch2023also,yang2023gd} have primarily targeted single-modal perception models, facilitating label-efficient transfer for either camera or LiDAR inputs (see Fig.\ref{fig:motivation}a). 
However, few studies\cite{calico} have explored pre-training methods for multi-modal perception models~\cite{liu2022bevfusion}. Moreover, most existing approaches~\cite{bevdistill,li2022simipu,calico} for multi-modal models often employ distinct optimization strategies for each modality, limiting their ability to fully leverage the complementary nature of multi-modal data.
Thus, we aim to develop a unified pretraining method that facilitates efficient fine-tuning for multi-modal perception models (see Fig.\ref{fig:motivation}b). To achieve this, we address two key challenges: \textbf{1}) learning transferable multi-modal representations, and \textbf{2}) unifying the optimization formulation across modalities.

On one hand, to foster self-supervised learning of multi-modal representations, we draw inspiration from Masked Auto Encoder (MAE)~\cite{he2022masked}. MAE’s mask-then-reconstruct approach has proven highly effective in learning robust representations. By requiring models to infer missing data from partial observations, it enhances generalization. This process encourages the network to capture high-level features that are shared across both appearance and geometric modalities, making it an excellent fit for transfer learning tasks. Recently, MAE-style pre-training has proven to be highly effective for single-modal  models~\cite{lin2024bev,voxel-mae,voxel-mae-wacv}. In this work, we extend MAE to learn transferable multi-modal representations for autonomous driving~\cite{liu2022bevfusion}, where the task of predicting missing information from both sparse LiDAR and (multi-view) camera data promotes the network's ability to generalize across different downstream tasks.

On the other hand, to achieve unified optimization of multi-modal representations, we incorporate the neural rendering mechanism of NeRF~\cite{aliev2020neural}. NeRF offers an elegant and physically interpretable solution for encoding scene characteristics, such as color and geometry, through differentiable rendering. We leverage NeRF in multi-modal perception pre-training for two key reasons: 1) radiance field rendering effectively mimics the imaging processes of both optical systems like LiDAR and cameras, enabling the model to jointly learn spatial and semantic representations; and 2) neural rendering unifies multi-modal reconstruction through a coherent physical model that seamlessly integrates information across modalities. By aligning LiDAR’s geometric precision with the camera’s rich visual details, neural rendering enables the extraction of shared multi-modal features that are both spatially coherent and semantically rich.

\begin{figure}[t!]
    \centering
    \includegraphics[width=0.48\textwidth]{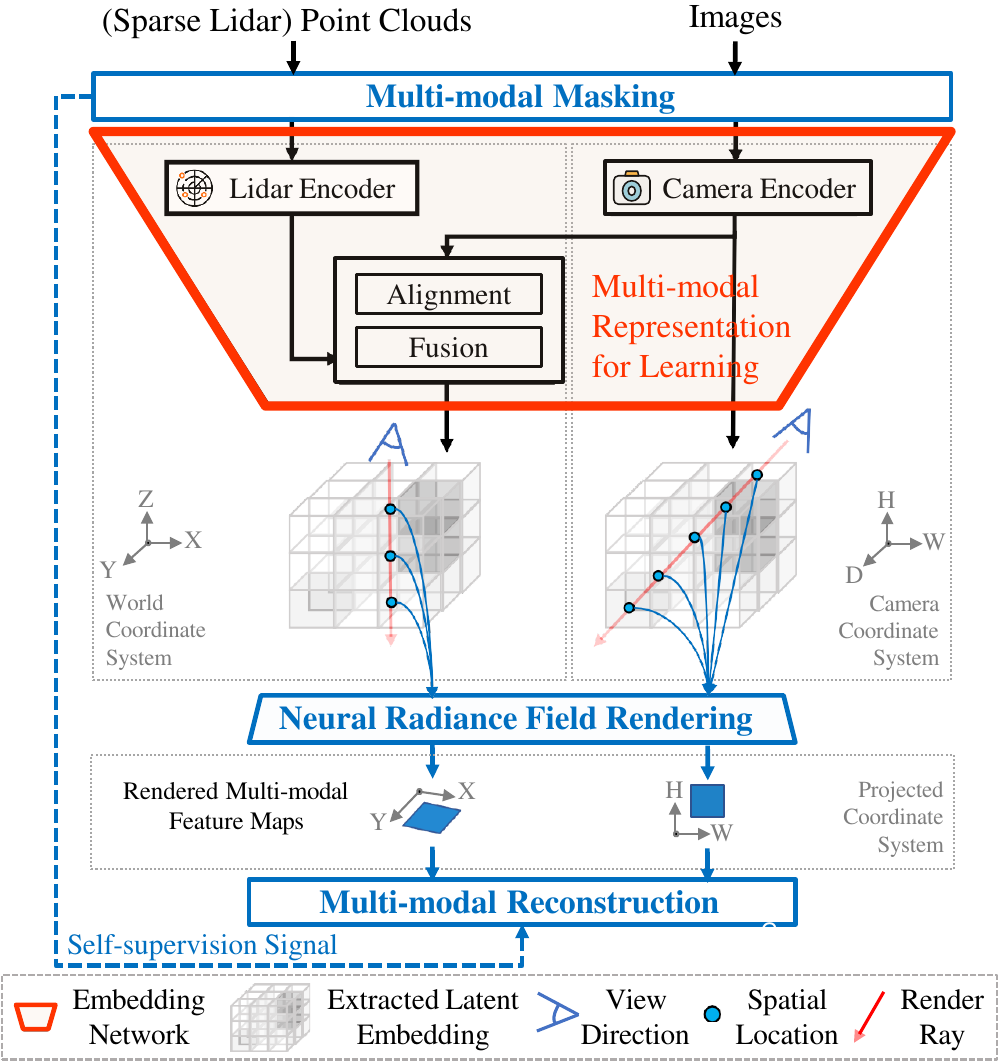}     \vspace{-6mm}
    \caption{\textbf{{Overview} of the NS-MAE pre-training framework.} Multi-modal representations are learned through self-supervised reconstruction of masked multi-modal inputs, using a differentiable neural rendering process.}
    \label{fig:teaser}
    \vspace{-2mm}
\end{figure}

Building on the synergy between MAE-style masked optimization and NeRF-style differentiable rendering, we propose \name{}, a unified self-supervised multi-modal perception pre-training framework that effectively learns transferable representations. As illustrated in Fig.\ref{fig:teaser}, \name{} integrates plug-and-play pre-training modules (blue blocks in Fig.\ref{fig:teaser}) with the embedding network of multi-modal perception models (black blocks in Fig.\ref{fig:teaser}). During pre-training, we apply modality-specific masking to both images and LiDAR point clouds separately, corrupting the input data to prevent trivial solutions and enhance the model’s ability to infer missing information. The multi-modal embedding network processes the corrupted modalities to extract embeddings. The embeddings are fused in two essential coordinate systems—world and camera coordinates—commonly used in perception models\cite{li2022bevformer}. For each view direction and spatial location, the embeddings are rendered into projected modality feature maps using differentiable volume rendering. These rendered feature maps are then supervised by the original images and point clouds through reconstruction-based self-supervised optimization.
We conduct extensive experiments to validate the transferability of the multi-modal representations learned through \name{} across diverse 3D perception models. 

The main contributions of this paper are: 
\begin{itemize}[leftmargin=*] 
\item We propose a self-supervised pre-training framework, NS-MAE, for LiDAR-Camera 3D perception, which unifies the optimization of multi-modal representations.
\item We incorporate plug-and-play components inspired by masked reconstruction and NeRF-like rendering, facilitating self-supervised learning and transferable multi-modal representation optimization.
\item We demonstrate the effectiveness of \name{} for both multi-modal and single-modal perception models, validating the transferability of multi-modal representations across 3D perception tasks with varying fine-tuning data levels. Notably, \name{} achieves superior BEV map segmentation performance than prior SOTA methods under the label-efficient fine-tuning scenarios.
\end{itemize}

\section{Related Work}
\label{sec:related_work}
\mypara{Masked AutoEncoder (MAE).}  MAE~\cite{he2022masked} demonstrates that directly reconstructing masked patches in raw pixel space can lead to good transferability and scalability for visual perception. Subsequent works perform reconstruction in a high-level feature space~\cite{zhou2021ibot,fang2022data} or handcrafted feature space~\cite{wei2022masked}. NeRF-MAE~\cite{irshad2024nerf} is a recent approach that employs masked reconstruction on NeRF inputs, using posed RGB images as supervision to enable generalizable NeRF representation learning.  Enlightened by these methods, we  tap the potential of introducing MAE from single-modal~\cite{bevdistill,lin2024bev} to multi-modal representation learning for 3D  perception in autonomous driving. 

\mypara{Neural Radiance Field (NeRF)}. NeRF~\cite{mildenhall2020nerf} shows impressive view
synthesis results by using implicit functions to encode volumetric density and color observations. To improve the few-shot generalization ability, data-driven priors recovered from general training data~\cite{yu2021pixelnerf} and mixed priors~\cite{chen2021mvsnerf} are leveraged to fill in missing information of test scenes. For efficient training, some works~\cite{wei2021nerfingmvs,deng2022depth} attempt to regulate the 3D geometry with depth. The neat formulation of the neural rendering mechanism is the basis for the optimization of both the appearance and geometry. Recently, Ponder~\cite{huang2023ponder} shows that NeRF-like pre-training head benefits indoor 3D  perception models with point cloud backbone and dense RGB-D input, our work differentiates by targeting outdoor driving perception models, utilizing sparse, weakly-aligned LiDAR data and RGB images. 

\mypara{LiDAR-Camera-based Driving Perception models.} \textbf{1}) {{Camera-only}} models~\cite{reading2021categorical, wang2021depth} focus on monocular 3D detection~\cite{kitti} and have evolved with larger benchmarks~\cite{Caesar2020nuScenesAM, waymo2020} to include range-view images~\cite{wang2022detr3d} and bird's-eye view (BEV) perception~\cite{huang2021bevdet, harley2023simple,li2022bevformer,han2024exploring,kim2024broadbev}. \textbf{2}) {{LiDAR-only}} models initially used raw point clouds~\cite{Yang20203DSSDP3, li2021lidar} or voxel/pillar representations~\cite{Zhou2018VoxelNetEL, Lang2019PointPillarsFE}, later integrating both~\cite{Zhou2019EndtoEndMF, Shi2020PVRCNNPF}. \textbf{3}) {{Multi-modal}} models~\cite{bai2022transfusion, li2022deepfusion, yang2022deepinteraction,liu2022bevfusion, liang2022bevfusion} leverage multiple modalities for more robust 3D perception.  Although pre-training has been explored for single-modal perception~\cite{li2022simipu,voxel-mae,xu2023mv,hess2023masked,krispel2024maeli}, unified multi-modal pre-training remains largely unexplored~\cite{calico}. Existing methods design separate strategies~\cite{li2022simipu,bevdistill,lin2024bev,calico} for self-supervised learning with RGB and depth, which limited the ability to capture shared features across modalities in a unified way. This motivates our exploration of more unified representation learning for multi-modal driving perception models.

\section{Approach}
\label{sec:method}

\subsection{Problem Formulation of LiDAR-Camera Pre-training}

 Given a tuple of images $\mathbf{I}=({I}_1, {I}_2, ...,{I}_N\in {\mathbb{R}}^{H\times W \times 3})$ collected from $N$ views, along with their corresponding camera parameters  $(\mathbf{P}_1\mathbf{K}_1, \mathbf{P}_2\mathbf{K}_2, ..., \mathbf{P}_N\mathbf{K}_N)$ (where $\mathbf{P}$ and $\mathbf{K}$ denote camera pose and intrinsic matrix) and LiDAR point clouds $\mathcal{P}=\{{p}_i=[x_i,y_i,z_i,r_i]\in \mathbb{R}^4 \}_{i=1,...,t}$,\footnote{$x,y,z$ denotes the position of a point in the world coordinate space; LiDAR intensity $r$ is optional for the input but is used as common practice for the LiDAR-based encoders~\cite{Zhou2018VoxelNetEL}} the goal of perception pretraining is to design a proxy task to learn the parameter set, \textit{i.e.}, transferable representation, of an embedding network $\mathcal{\phi}_{emb}$, which can be used to initialize the parameter set of the downstream 3D perception model $\mathcal{\phi}_{down}(\supset \mathcal{\phi}_{emb}$) for further fine-tuning.

\subsection{Pipeline Overview}

As shown in Fig.~\ref{fig:teaser}, the NS-MAE pre-training pipeline involves three key steps for learning transferable multi-modal representations. First, the input images and voxelized LiDAR point clouds are masked independently (Sec.~\ref{Sec: Multi-modal Masking}). Next, the masked embeddings are rendered into color and depth maps using neural rendering from two viewpoints: perspective and bird’s-eye views (Sec.~\ref{Sec: Radiance Field Rendering}, Sec.~\ref{Sec:Rendering Viewpoint Selection}, Sec.~\ref{Sec: Radiance Field Rendering Target}). Finally, the rendered outputs are optimized against ground-truth images and point clouds through self-supervised multi-modal reconstruction (Sec.~\ref{Sec: Multi-modal Reconstruction}).

\subsection{Training Architecture}
\label{Sec: Architecture Details}
The pre-training architecture, a typical embedding network for multi-modal perception models, comprises the following components (see Fig.~\ref{fig:teaser}):

\mypara{Camera encoder.} The camera encoder processes masked images and generates the perspective-view image embedding $\mathbf{e}_{I}^{C}$ in camera coordinates.

\mypara{LiDAR encoder.} The LiDAR encoder processes masked LiDAR voxels and produces the LiDAR embedding $\mathbf{e}_{L}^{W} \in {\mathbb{R}}^{X \times Y \times Z \times C_L}$ in world coordinates.

\mypara{Alignment module.} This module aligns the camera and LiDAR embeddings by transforming the perspective-view image embedding $\mathbf{e}_{I}^{C}$ from camera coordinates into the world coordinate system, resulting in $\mathbf{e}_{I}^{W} \in {\mathbb{R}}^{X \times Y \times Z \times C_I}$. This alignment enables both modalities to operate in a unified world space, improving integration.

\mypara{Fusion block.} The fusion block combines the aligned camera embedding $\mathbf{e}_{I}^{W}$ with the LiDAR embedding $\mathbf{e}_{L}^{W}$ through simple concatenation $[\mathbf{e}_{I}^{W};\mathbf{e}_{L}^{W}] \in {\mathbb{R}}^{X \times Y \times Z \times (C_I + C_L)}$. This fused multi-modal representation allows the mutual enhancement of spatial and semantic information from both modalities, leading to more robust multi-modal perception.

\subsection{Multi-modal Masking}
\label{Sec: Multi-modal Masking}

\noindent\textbf{Motivation.} Masking prevents the network from learning trivial identity mappings~\cite{he2022masked} by enforcing generative reconstruction of missing regions, thereby learning useful and generalizable representations.

\mypara{Image Masking.}  The original unmasked image $I$ is first divided into regular non-overlapping image patches. Then, a random binary mask $M\in {\{0,1\}}^{H\times W}$ is applied to mask out a large portion (50\%) of image patches by replacing them with a \texttt{[MASK]} token ($\in \mathbb{R}^{s\times s\times 3}$, where $s\times s$ denotes the patch size). Afterward, the image that is partially masked out is sent to the camera encoder for embedding extraction.

\mypara{LiDAR Masking.} After voxelizing the input LiDAR point cloud, we mask a large fraction of non-empty voxels (70\% to 90\%). Then, the partially-masked voxels are sent in the LiDAR encoder to generate its embedding.

\subsection{NeRF-like Rendering Network}
\label{Sec: Radiance Field Rendering}

\noindent\textbf{Motivation.} By leveraging NeRF’s ability to jointly model appearance and geometry, we enable the network to learn structured representations that integrate spatial and semantic information from both LiDAR and camera data.

\mypara{Vanilla rendering mechanism}.  
In NeRF, the rendering network $f$ takes as input a 3D point $\textbf{x} \in \mathbb{R}^3$ (in world coordinates) and a viewing direction ${\omega} \in \mathbb{S}^2$ (spherical coordinate system). It outputs the voxel density $\sigma\in \mathbb{R}$, which encodes the local geometry, and the RGB color $\textbf{c}\in \mathbb{R}^3$, representing the scene's appearance. This coupling of appearance and geometry allows for efficient view rendering.

\mypara{Conditional rendering network for pre-training.} As our goal (representation learning) is different from the goal of vanilla NeRF (view synthesis), we leverage a reformulated rendering network design. In particular, we additionally introduce a latent multi-modal embedding $\mathbf{e}$ from the embedding network of the perception model to the inputs of the rendering network $f$, like so: $f(\textbf{x}, \omega, \mathbf{e}) = (\sigma, \textbf{c})$. Thus, the gradient from differential rendering and the reconstruction-based self-supervision stages can be back-propagated to the embedding network for end-to-end learning.

\subsection{NeRF-like Rendering Viewpoint Selection}
\label{Sec:Rendering Viewpoint Selection}

\mypara{Viewpoint selection strategy}.
Our pre-training strategy leverages two key viewpoints—bird's-eye view (BEV) and perspective view (PER)—to capture complementary aspects of 3D perception. BEV provides a top-down view for spatial context and geometry, while PER captures object-level details from the camera's perspective. Combining these viewpoints enhances multi-modal representation learning, leading to improved performance across 3D perception tasks, as shown in our ablation studies (see Section~\ref{ablation_study}).

\subsection{NeRF-like Rendering Target}
\label{Sec: Radiance Field Rendering Target}

\noindent\textbf{Motivation.} Our rendering targets, including both color and depth, ensure that the network learns modality-agnostic representations, making it robust to various input types and enhancing transferability to downstream tasks.

\mypara{Vanilla Color Rendering}.
Given the pose $\mathbf{P}$ and intrinsic $\mathbf{K}$ of a \textit{virtual camera}, we shoot rays $\mathbf{r}(t) = \mathbf{o} + t\mathbf{d}$ originating from the $\mathbf{P}$’s center of projection $\mathbf{o}$ in direction $\omega$ derived from its intrinsic $\mathbf{K}$ to render the RGB color $\hat{\mathbf{C}}(\mathbf{r})$ via standard volume rendering \cite{kajiya1984ray}, which is formulated as:
\begin{align}
 \hat{\mathbf{C}}(t) = \int_{0}^{\infty} T(t)\sigma(t)\mathbf{c}(t) dt,
\label{equ:render_color_c}
\end{align}
where $\mathbf{c}(t)$ and $\sigma(t)$ are the differential color radiance and density, and $T(t) = \exp(-\int_{0}^t \sigma(s) ds)$ checks for occlusions by integrating the differential density from $0$ to $t$. Specifically, the discrete form can be approximated as:
\begin{align}
    \hat{\mathbf{C}}(\mathbf{r}) = \sum_{i=1}^{N}T_{i}(1 - \mathrm{exp}(-\sigma_{i}\delta_{i}))\mathbf{c}_{i}, 
\label{equ:render_color_d}
\end{align}
where $N$ is the number of sampled points along the ray, $\delta_{i} = t_{i+1}-t_{i}$ is the distance between two adjacent ray samples and the accumulated transmittance $T_{i}$ is $\mathrm{exp}(-\sum_{j=1}^{i-1}\sigma_{j}\delta_{j})$.

\mypara{Multi-modal rendering for pre-training}. For multi-modal pre-training, the rendering targets can be extended to unleash the power of multi-modal data. In particular, apart from the color that reflects the semantics of the scene, the LiDAR ray that captures 3D geometry in the form of point clouds is also a kind of radiance. 
Going beyond the differential RGB color radiance $\mathbf{c}(t)$ for color rendering in Eq.~(\ref{equ:render_color_c}), we leverage the differential radiance of a kind of \textit{modality} $\mathbf{a}(t)$ for multi-modal rendering. Specifically, the rendering process to derive the projected feature map $\hat{\mathbf{A}}(t)$ can be formulated as:
 \begin{align} 
 \hat{\mathbf{A}}(t) = \int_{0}^{\infty} T(t)\sigma(t)\mathbf{a}(t) dt,
\label{equ:depth_render_m}
\end{align}
\noindent Typically, to render the projected 3D point cloud feature map, \textit{i.e.}, 2D depth $\hat{\mathbf{D}}(\mathbf{r})$, the discrete form can be expressed as:
\begin{align}
    \hat{\mathbf{D}}(\mathbf{r}) = \sum_{i=1}^{N}(T_{i}(1 - \mathrm{exp}(-\sigma_{i}\delta_{i}))\sum_{j=1}^{i-1}\delta_j), 
\label{equ:depth_render_d}
\end{align}

\subsection{Multi-modal Reconstruction-based Optimization}
\label{Sec: Multi-modal Reconstruction}

\noindent\textbf{Motivation.} Reconstruction-based optimization with multi-modal data ensures alignment between outputs and original sensor data, reinforcing consistency across modalities and improving multi-modal fusion during pre-training.

\mypara{Vanilla color reconstruction objective.} Given a set of rendering rays $\mathcal{S}_r$ passing through the pixels of the original image, the goal of this step is to minimize the square of the $L_2$-norm of the difference between the ground truth color $\mathbf{C}(r)$ and the rendered color $\hat{\mathbf{C}}(r)$ of ray $r$:
\begin{align}
    \mathcal{L}_{\mathrm{C}} = \frac{1}{|\mathcal{S}_r|}\sum_{r \in \mathcal{S}_r}\| \hat{\mathbf{C}}(\mathbf{r}) - \mathbf{C}(\mathbf{r}) \|_{2}^2,
    \label{equ:loss_c}
\end{align}
\mypara{Multi-modal reconstruction objective for pre-training.} Given a set of rendering rays $\mathcal{S}_r$ passing through the ground-truth target, we minimize the $L_p$-norm of the difference between the ground-truth view-specific projected feature map $\mathbf{A}(r)$ and the rendered result $\hat{\mathbf{A}}(r)$ of ray $r$ to the $p$-th power:
\begin{align}
    \mathcal{L}_{\mathrm{A}} = \frac{1}{|\mathcal{S}_r|}\sum_{r \in \mathcal{S}_r}\| \hat{\mathbf{A}}(\mathbf{r}) - \mathbf{A}(\mathbf{r}) \|_{p}^p,
    \label{equ:loss_mm}
\end{align}
\noindent The {overall objective function}  jointly optimize the reconstruction for multiple ($K$) view-specific modalities:
\begin{align}
\mathcal{L} =  \sum_{k=1}^{K} (\lambda_{k} \cdot \mathcal{L}_{\mathrm{A}_k} ) \label{equ:loss_final}
\end{align}
 $\lambda_{k}$ indicates the coefficient to modulate the $k$-th sub-loss.

\renewcommand{\arraystretch}{0.6}
\begin{table}[t!]
\centering
\caption{mIoU Performance Comparison on BEV Map Segmentation with Varying Fine-Tuning Data Ratios for nuScenes~\cite{nuscenes}.}
\vspace{-2mm}
\resizebox{0.48\textwidth}{!}
	{   \setlength{\tabcolsep}{3pt}
\begin{tabular}{l|c|cc}
\toprule
\multirow{2}{*}{Pretraining Method} & {Pretraining Strategy} &\multicolumn{2}{c}{Sampling Ratio} \\ \cmidrule{3-4}
 &{For Each Modality}  & {10\%}  & {50\%} \\
\midrule
Rand. Init.  & No & 43.8 & 55.4 \\
+ {PRC~\cite{calico}} & Separate (Contra.) & 45.3 & 56.0 \\
+ SimIPU~\cite{li2021simipu} & Separate (Contra.) & 45.1  & 55.9 \\
+ 
 {PRC~\cite{calico}+BEVDistill}~\cite{bevdistill}&Separate (Contra.\&Distill) & 46.4  & 56.4 \\
+ {CALICO}~\cite{calico} &Separate (Contra.\&Distill) & {47.3} & {56.7} \\
  \rowcolor{place3d_blue!15}   + {NS-MAE} &  Unified (Mask+Render) & \textbf{47.7} & \textbf{56.9} \\
\bottomrule
\multicolumn{4}{l}{\scriptsize{`Contra.' denotes contrastive learning. `Distill' denotes distillation. }}\\
\multicolumn{4}{l}{\scriptsize{`Mask+Render' denotes masked rendering-based optimization. }}\\
\end{tabular}} \vspace{-2mm}
\label{tab:comparison_with_prior}
\end{table}

\renewcommand{\arraystretch}{0.8}
\begin{table}[t]

\caption{Comparison to supervised pre-training for BEVFusion~\cite{liu2022bevfusion} on 3D object detection on nuScenes \cite{nuscenes} $val$. 
 	}
\vspace{-2mm}
\resizebox{0.48\textwidth}{!}
	{
        \setlength{\tabcolsep}{2pt}
	\centering
	\begin{tabular}{l|c|cc|cc}
          \toprule
        \multirow{2}{*}{Method}  & \multirow{2}{*}{Pre-training}  &\multicolumn{2}{c|}{{Image: $128\times352$}}  &\multicolumn{2}{c}{{Image: $256\times704$}}     \\
         & \multirow{2}{*}{Type} &\multicolumn{2}{c|}{{\#Lidar Sweep: 1}}  &\multicolumn{2}{c}{{\#Lidar Sweep: 9}}     \\ \cmidrule(lr){3-4} \cmidrule(lr){5-6} 
       (L: LiDAR; C: Camera)   & & \quad mAP & NDS & \quad mAP & NDS  
         \\  \midrule
          Rand. Init. (L+C) & No   & \quad{50.5} &{53.3} &\quad {60.8} &{64.1}  \\ 
          + iNet (C)  &Fully-sup.  &\quad{51.4} &{54.1}      &\quad{62.8} &{65.3}     
         \\  
  \rowcolor{place3d_blue!15}        + \name{}  (L+C)  & Self-sup.  & \quad\textbf{{51.5}} &\textbf{54.7}      & \quad\textbf{{63.0}} &\textbf{65.5} 
         \\          

        \bottomrule 
        \multicolumn{6}{l}{\scriptsize{+ iNet indicates pre-training on ImageNet~\cite{deng2009imagenet} in a supervised way. }}\\
	\end{tabular}
       }
        \vspace{-4mm}
	\label{tab:nuscene_det_sup_train}

\end{table}

\renewcommand{\arraystretch}{0.5}
\begin{table*}[t!]
\small
\centering
\caption{3D object detection results for Multi-modal Perception model (BEVFusion~\cite{liu2022bevfusion}) on \nus{} \cite{Caesar2020nuScenesAM} $val$.}
\vspace{-2mm}
\resizebox{1.0\textwidth}{!}
{
\setlength{\tabcolsep}{6pt}
\centering
\begin{tabular}{l|ccc|cccccccccc|cc}
\toprule
\multirow{2}{*}{Method} & \multirow{2}{*}{Modality} & \multirow{2}{*}{\#Sweep} & \multirow{2}{*}{\#ImgSize}  & \multicolumn{10}{c|}{Per-class mAP} & \multirow{2}{*}{{\small{mAP}}} & \multirow{2}{*}{{\small{NDS}}} \\ \cmidrule(lr){5-14}
&  &  &  & \small{Car} & \small{Truck} & \small{C.V.} & \small{Bus} & \small{Trail.} & \small{Barr.} & \small{Moto.} & \small{Bike} & \small{Ped.} & \small{T.C.}  &  & \\
\midrule
BEVFusion~\cite{liu2022bevfusion}& {L+C} &{1} &{$128\times 352$}   & 81.1 & {37.4} & 12.3 & 59.0 & \textbf{31.5} & 64.1 & {46.7} & {28.9} & {80.3} & {63.1}  & {50.5} &{53.3}\\ 
\rowcolor{place3d_blue!15}      + \name & {L+C} &{1} &{$128\times 352$}  & \textbf{{81.6}} & \textbf{{40.1}} & \textbf{{13.9}}& \textbf{{59.8}} & {30.1} & \textbf{{64.7}} & {\textbf{{48.9}}} & {\textbf{{30.3}}} & {\textbf{{81.0}}} & \textbf{{64.4}}    & \textbf{{{51.5}}} &\textbf{{54.7}}       
\\         
\midrule 
\textcolor{black}{BEVFusion}~\cite{liu2022bevfusion} & {L+C} &{9} &{$256\times 704$}  & 87.4 & {40.4} & {\textbf{{25.7}}} & 67.0 & \textbf{{38.8}} & 71.6 & 68.2 & {48.6} & {{85.5}} & {{74.5}} &  {60.8} &{64.1} \\
\rowcolor{place3d_blue!15}         + {\name} & {L+C} &{9} &{$256\times 704$}    & {\textbf{{88.1}}} & {\textbf{{45.9}}} & {25.1}& {\textbf{{68.8}}} & {{37.2}} & \textbf{{73.8}} & \textbf{{70.8}} & \textbf{{56.6}} & \textbf{{86.9}} & \textbf{{77.4}} & \textbf{{63.0}} &\textbf{{65.5}}  \\ 

\bottomrule
\end{tabular}
}

\vspace{-4mm}
\label{tab:nuscene_det}
\end{table*}

\begin{table}[t]
  \caption{3D object detection results on KITTI-3D~\cite{kitti} $val$.}
  \vspace{-2mm}
  \centering
  \setlength{\tabcolsep}{2.4pt}
  \resizebox{0.48\textwidth}{!}{
  \begin{tabular}{l|c|ccc|ccc}
    \toprule
    \multirow{2}{*}{Method} &\multirow{2}{*}{$DB$} & \multicolumn{3}{c|}{AP${}_{BEV}$} & \multicolumn{3}{c}{AP${}_{3D}$}\\ \cmidrule(lr){3-5} \cmidrule(lr){6-8}  
    {} && Easy & Moderate & Hard & Easy & Moderate & Hard \\
    \midrule
    VFF-SECOND & &  {91.71} &{85.77} & {83.54} & {87.25} & {76.44} & {74.02} \\
    \rowcolor{place3d_blue!15}         + \name & &  \textbf{92.65} & \textbf{88.24} & \textbf{85.83} & \textbf{88.25} & \textbf{78.40} & \textbf{74.37} \\
    \midrule
     VFF-SECOND &$\checkmark$ &  {92.83} &{88.92} & {88.22} & {89.45} & {82.32} & {79.39} \\   
  \rowcolor{place3d_blue!15}       + \name  &$\checkmark$&  \textbf{93.04} & \textbf{90.43} & \textbf{88.46} & \textbf{91.48} & \textbf{82.58} & \textbf{79.77} \\  
  \midrule 
     VFF-PVRCNN   &$\checkmark$&  {92.65} &{90.86} & {88.55} & {91.75} & {85.09} & {82.68} \\  
  \rowcolor{place3d_blue!15}         + \name &$\checkmark$ &  \textbf{92.94} & \textbf{91.00} & \textbf{90.49} & \textbf{92.03} & \textbf{85.31} & \textbf{83.05} \\   \midrule
     VFF-VoxelRCNN   &$\checkmark$&  \textbf{95.67} &{91.56} & {89.20} & {92.46} & {85.25} & {82.93} \\  
    \rowcolor{place3d_blue!15}       + \name &$\checkmark$ &  {95.57} & \textbf{91.69} & \textbf{89.23} & \textbf{92.51} & \textbf{85.59} & \textbf{82.95} \\   
    \bottomrule
    \multicolumn{7}{l}{\scriptsize{$DB$ indicates database-sampler~\cite{Zhu2019ClassbalancedGA} data augmentation is used.}}\\
    \end{tabular}
    }
    \label{tab:kitti}
    \vspace{-2mm}
\end{table}

\renewcommand{\arraystretch}{0.5}
\begin{table*}[t!]
\small
\caption{{3D object detection results for Camera-only (Top) and LiDAR-only (Down) Perception} models on \nus{} \cite{Caesar2020nuScenesAM} $val$. \\ ${2\times}$ denotes doubled training epochs. \#Sweep denotes the LiDAR sweep number. \#ImgSize denotes the image size. }
\vspace{-2mm}
\resizebox{\textwidth}{!}
	{
        \setlength{\tabcolsep}{7pt}
	\centering
	\begin{tabular}{l|ccc|cccccccccc|cc}
          \toprule
         \multirow{2}{*}{Method} & \multirow{2}{*}{Modality} & \multirow{2}{*}{\#Sweep} & \multirow{2}{*}{\#ImgSize}  & \multicolumn{10}{c|}{Per-class mAP} & \multirow{2}{*}{{\small{mAP}}} & \multirow{2}{*}{{\small{NDS}}} \\ \cmidrule(lr){5-14}
          &  &  &  & \small{Car} & \small{Truck} & \small{C.V.} & \small{Bus} & \small{Trail.} & \small{Barr.} & \small{Moto.} & \small{Bike} & \small{Ped.} & \small{T.C.}  &  & \\
         \midrule
          BEVDet~\cite{huang2021bevdet}& C &/ & $256\times 704$ & 35.3 & {15.7} & {2.7} & \textbf{{18.2}} & 5.5 & 31.0 & 19.8 & 18.3 & 26.9 & 46.0  & {21.9} &{29.4}\\ 
      \rowcolor{place3d_blue!15}         + \textcolor{black}{\name}  &C &/ & $256\times 704$     &{\textbf{{37.1}}} &{\textbf{{18.0}}} &\textbf{{3.6}} &{{18.0}} &\textbf{{7.0}}  &\textbf{{{41.3}}} &\textbf{{{20.3}}} &\textbf{{{19.8}}} &\textbf{{28.1}} &\textbf{{{47.6}}}        &\textbf{{24.1}} &\textbf{{32.1}}       
         \\  \midrule 
            BEVDet${}^{2\times}$~\cite{huang2021bevdet}& C &/ & $256\times 704$ & 37.1 & {16.4} & {2.7} & \textbf{{18.9}} & 5.8 & 42.0 & 20.7 & 18.3 & 28.3 & 49.0 & {23.9} &{31.8}\\ 
    \rowcolor{place3d_blue!15}           + \textcolor{black}{\name}  &C &/ & $256\times 704$   &\textbf{{{38.3}}} &\textbf{{{17.9}}}&\textbf{{4.5}} &{{{18.3}}} &\textbf{{8.1}}  &\textbf{{{47.1}}}  &\textbf{{{21.5}}} &\textbf{{{18.8}}} &\textbf{{29.4}} &\textbf{{{49.8}}}   &\textbf{{25.4}} &\textbf{{33.9}}       
         \\ \midrule
 
          CenterPoint~\cite{Yin2020Centerbased3O}& L &9 & /    & 80.9 & 52.4 & 14.4 & \textbf{{64.0}} & 29.6 & 58.7 & 59.4 & 45.6 & 80.4 & 60.8  & {54.6} &{61.3}\\ 
          + {VoxelMAE}~\cite{voxel-mae}  &L &9 & /      &{{80.6}} &\textbf{{{53.7}}} &{13.7} &{{63.2}} &{29.2}  &\textbf{{{61.1}}}  &\textbf{{{60.5}}} &{{45.4}} &{80.4} &{{61.1}}  &{{54.9}} &{{61.4}}\\
   \rowcolor{place3d_blue!15}            + {\name}  &L &9 & /      &\textbf{{{81.2}}} &{{53.0}} &\textbf{{{14.7}}}&{{{63.7}}} &\textbf{{{30.2}}} &{{60.0}}  &{{60.1}} &\textbf{{{47.1}}} &\textbf{{{81.6}}} &\textbf{{{61.3}}}      &\textbf{{55.3}} &\textbf{{62.1}}    \\ 
        \bottomrule
	\end{tabular}
       }
        \vspace{-2mm}
        \centering

\vspace{-3mm}
	\label{tab:nuscene_det_merged}
\end{table*}
\begin{table}[t!]
        \caption{BEV map segmentation results for Multi-modal and Camera-only methods on \nus{} \cite{Caesar2020nuScenesAM} $val$.}
\vspace{-2mm}
\small\centering
\setlength{\tabcolsep}{0.7mm}
\label{tab:results:segmentation}
\resizebox{0.48\textwidth}{!}
	{
        \begin{tabular}{l|c|cccccc|c}
            \toprule
            \multirow{2}{*}{Method} & \multirow{2}{*}{Modality} & \multicolumn{6}{c|}{Per-class IoU} & \multirow{2}{*}{mIoU} \\   \cmidrule(lr){3-8} 
             &  & Dri. & P.C. & Walk. & S.L. & Car. & Div. & \\
            \midrule
             {BEVFusion}~\cite{liu2022bevfusion}& {L+C} & {75.0} & {42.6} & {52.6} & {24.4} & {26.6} & {36.0} & \textcolor{black}{42.9} \\           
   \rowcolor{place3d_blue!15}              + \name & L+C & \textbf{78.0} & \textbf{45.9} & \textbf{55.5} & \textbf{26.1} & \textbf{35.4} & \textbf{38.9} & \textcolor{black}{\textbf{46.6}} \\ \midrule 

          {BEVDet}~\cite{huang2021bevdet}& C & {72.7} & {35.6} & {44.7} & {21.1} & {34.0} & {32.3} & {40.1} \\    
     \rowcolor{place3d_blue!15}             + \name &  C& \textbf{76.1} & \textbf{39.9} & \textbf{49.0} & \textbf{23.5} & \textbf{41.6} & \textbf{35.6} & \textbf{44.3} \\
            \bottomrule
        \end{tabular}
        }
        \label{tab:nuscene_seg}
\vspace{-2mm}
\end{table}

\section{Experiments}
\label{sec:experiment}

\subsection{Pre-training Setup and Details}
\label{Sec: Pre-training Setup}

\mypara{Training strategy.} The embedding network is pre-trained for 50 epochs using the AdamW optimizer~\cite{adamw} with a learning rate of 1e-4 and weight decay of 0.01. A one-cycle learning rate scheduler~\cite{one-cycle} is employed. Training is performed on 8 NVIDIA V100 GPUs with a batch size of 16.

\mypara{Masking.} For images, masking patch sizes are $4\times4$ and $8\times8$ for resolutions $128\times352$ and $256\times704$, with a 50\% masking ratio. For LiDAR, the point cloud range is set to -54$\sim$54 (m) for $X$ and $Y$, and -5$\sim$3 (m) for $Z$. The voxel size is [0.075, 0.075, 0.2] (m), with a 90\% masking ratio for non-empty voxels, or in a range-aware manner~\cite{voxel-mae}.

\mypara{Rendering.} The rendering network uses convolutional layers to transform embeddings into sigma-field and color feature volumes. For BEV and perspective views, the rendering parameters $\delta$ are set to 0.2 and 0.8, respectively.

\mypara{Reconstruction.} The reconstruction targets include color field maps ($\mathbf{C}$) for camera images, perspective-view depth ($\mathbf{D}^{PER}$) projected from LiDAR onto image planes, and BEV depth ($\mathbf{D}^{BEV}$) projected onto the BEV plane. Rays are emitted orthogonally to the BEV plane for depth rendering and pass through the image plane for color and depth rendering. The loss parameters are set as $p=2$ for color and $p=1$ for depth, with coefficients normalized at 1e4 and 1e-2 for color and depth, respectively.

\mypara{Implementation.} The network is implemented in PyTorch using the MMDetection3D framework~\cite{mmdet3d}. Standard LiDAR and image augmentations are applied, excluding those requiring ground-truth labels, such as database-sampling~\cite{Zhu2019ClassbalancedGA}.  Detailed experiment configurations, \textit{e.g.},  hyperparameters, are available
in our open-source repository for reproducibility.

\subsection{Dataset and Evaluation Metric}
\label{dataset_and_evaluation_metric}

\mypara{\nus{}}~\cite{Caesar2020nuScenesAM} is a large-scale 3D perception dataset with six surrounding-view cameras and LiDAR point clouds per frame. The evaluation metrics for 3D object detection and BEV map segmentation tasks are nuScenes detection score (NDS) and mean average precision (mAP), and Intersection-over-Union (IoU), respectively.

\mypara{KITTI-3D}~\cite{kittisplit} is a widely-used mall-scale benchmark for 3D object detection. Evaluation metrics include 3D IoU (AP${}_{3D}$) and BEV IoU (AP${}_{BEV}$) with average precision.

\subsection{Main Results}

\mypara{Comparisons with prior SOTA self-supervised pre-training methods.} 
As shown in Table~\ref{tab:comparison_with_prior}, our unified NS-MAE framework surpasses state-of-the-art methods~\cite{calico,li2022simipu,bevdistill} that use separate pre-training strategies. On the BEV map segmentation task in nuScenes~\cite{nuscenes}, NS-MAE consistently achieves higher accuracy across different label-efficient fine-tuning data ratios, demonstrating superior multi-modal representation learning.

\mypara{Comparisons to supervised camera-only pre-training.}
Table~\ref{tab:nuscene_det_sup_train} shows that NS-MAE outperforms conventional fully-supervised ImageNet-based pre-training~\cite{deng2009imagenet}, highlighting the superiority of multi-modal self-supervised pre-training for 3D perception tasks.

\mypara{Transferability under standard fine-tuning.}
\label{main_transfer_results}
Here, we evaluate the effectiveness of NS-MAE using 100\% labeled data for fine-tuning. We adhere to the baseline models' fine-tuning setups, training on \nus{} for 20 epochs and KITTI-3D for 80 epochs, without  CBGS~\cite{Zhu2019ClassbalancedGA} trick.

\begin{figure}[t]
    \centering
    \includegraphics[width=0.48\textwidth]{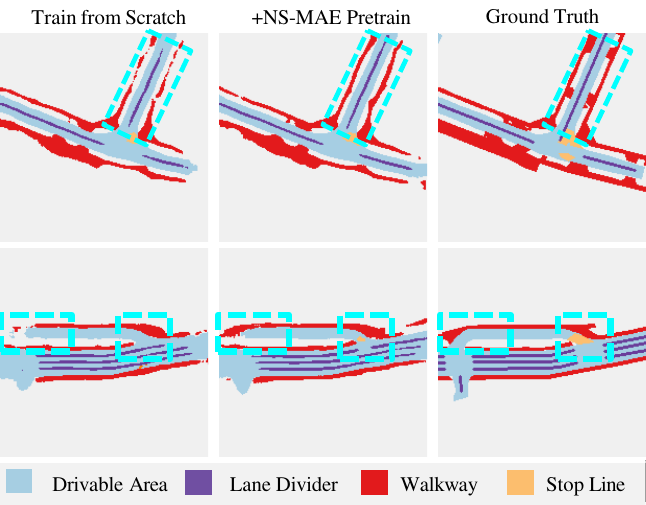} 
    \vspace{-6mm}
    \caption{{Qualitative comparisons of BEV map segmentation results} on \nus ~\cite{nuscenes} $val$ with multi-modal perception model BEVFusion~\cite{liu2022bevfusion}.} 
    \vspace{-2mm}
    \label{fig:bev_map_seg_result}
\end{figure}

\subsubsection{3D Object Detection Task}

Table~\ref{tab:nuscene_det} demonstrates that \name{} enhances the performance of the multi-modal BEVFusion model. Notably, for the multi-modal BEVFusion with multi-sweep LiDAR and higher image resolution, \name{} provides a notable boost, with a 2.2\% increase in mAP and a 1.4\% increase in NDS. Additionally, in Table~\ref{tab:kitti}, \name{} pre-training improves nearly all sub-metrics for the VFF~\cite{li2022voxel} baseline across various 3D detection heads.

In Table~\ref{tab:nuscene_det_merged}, we show the generality of pre-trained representation via \name{} to be transferred to single-modal, \textit{i.e.}, camera-only and LiDAR-only, models. For the camera-only model BEVDet~\cite{huang2021bevdet}, {{\name{}} {brings more than 2\% NDS improvement}} for the settings of both the default (20) and doubled ($2\times$: 40) training epochs. For the LiDAR-only CenterPoint~\cite{Yin2020Centerbased3O}, the transferable representation learned via our \name{} consistently shows its effectiveness. Moreover, compared to Voxel-MAE \cite{voxel-mae} that targets pre-training for the LiDAR-only perception models, our \name{} shows better performance (62.1\% \textit{v.s.} 61.4\% in NDS).

\subsubsection{BEV Map Segmentation Task}
NS-MAE significantly improves the performance, boosting mIoU by 3.7\% and 4.2\% for multi-modal and camera-only models (Table~\ref{tab:nuscene_seg}), with qualitative improvements shown in Fig.~\ref{fig:bev_map_seg_result}.

\renewcommand{\arraystretch}{0.5}
\begin{table}[t]
\small
\caption{Label-Efficient 3D object detection ({Top}) and BEV map segmentation ({Bottom}) on \nus{} \cite{Caesar2020nuScenesAM} $val$.}
\setlength{\tabcolsep}{8.6pt}
\centering
\vspace{-2mm}
\resizebox{0.48\textwidth}{!}
	{
\begin{tabular}{l|c|c|c|c|c}
\toprule
\multirow{2}{*}{Method} & \multirow{2}{*}{Metric} &\multicolumn{4}{c}{Sampling Ratio} \\
\cmidrule(lr){3-6} 
   &  & \multicolumn{1}{c}{  1\%} & \multicolumn{1}{c}{  5\%} & \multicolumn{1}{c}{10\%}  & \multicolumn{1}{c}{100\%}\\ \midrule
\midrule 
{BEVFusion~\cite{liu2022bevfusion}}&{mAP} 
              & {26.2}  & {46.1}& {54.2}  & \textcolor{black}{60.8} \\
  \rowcolor{place3d_blue!15}    {+ \name} &mAP& \textbf{{30.2}}  & \textbf{{47.6}}& \textbf{55.9}  & \textbf{{63.0}} \\
 \midrule 
{BEVFusion~\cite{liu2022bevfusion}}&{NDS} 
              & {44.2}  & {55.4}& {60.3}  & \textcolor{black}{64.1} \\
  \rowcolor{place3d_blue!15}   {+ \name} &NDS& \textbf{{45.4}}  & \textbf{{57.0}}& \textbf{61.4}  & \textbf{{65.5}} \\ \midrule \midrule
{BEVFusion~\cite{liu2022bevfusion}}&{mIoU} & {29.7}  & {39.4}& {41.3}  & \textcolor{black}{42.9} \\
  \rowcolor{place3d_blue!15}   \textcolor{black}{+ \name} & mIoU& \textbf{{31.1}}  & \textbf{{41.6}}& \textbf{{45.1}}  & \textcolor{black}{\textbf{{46.6}}} \\
\bottomrule
  \multicolumn{6}{l}{\scriptsize{Images of size $256\times704$ and multi-sweep (9) LiDAR points are used as input.}}\\
\end{tabular}
}

\label{tab:nuscene_label_efficient_det}
\end{table}

\begin{table}[t]
\small
\caption{{Ablation study} with the multi-modal perception baseline method BEVFusion~\cite{liu2022bevfusion} on \nus{} \cite{Caesar2020nuScenesAM} $val$ .}
\vspace{-2mm}
\setlength{\tabcolsep}{6.pt}
\centering
\resizebox{0.48\textwidth}{!}
{
\begin{tabular}{l|c|cccc|cc}
\toprule
\multirow{2}{*}{Setting}  & \multirow{2}{*}{Masking} &\multicolumn{4}{c|}{Rendering Targets} & \multirow{2}{*}{mAP} & \multirow{2}{*}{NDS}  \\ \cmidrule(lr){3-6} 
&     &  $\mathbf{C}$ & $\mathbf{D}^{PER}$& $\mathbf{D}^{BEV}$ & $\mathbf{D}^{Rand}$&   &  \\ \midrule
Baseline &           & &    &   &   &  {50.5} &{53.3}  \\ \midrule
(A1) &       &$\checkmark$&  &&  &{50.2}  &{53.9}\\     
(A2) &      $\checkmark$&$\checkmark$& & & &{50.7} &{54.2}     
\\   \midrule
(B0) &     $\checkmark$  &  &  &&$\checkmark$ & 50.7 & 53.6  \\
(B1) &     $\checkmark$  &  & $\checkmark$ & && {{50.9}} &{54.2} \\ 
(B2) &     $\checkmark$  &  &  &$\checkmark$& & 51.0 & 54.3  \\ 
(B3) &      &  &  $\checkmark$&$\checkmark$ && 49.6 & 52.8\\ 
(B4) &    $\checkmark$  &  &  $\checkmark$&$\checkmark$  & & {{51.3}} &{54.4}\\  \midrule
\rowcolor{place3d_blue!15}     Default &      $\checkmark$  &  $\checkmark$ &  $\checkmark$&$\checkmark$ & &\textbf{{51.5}} &\textbf{{54.7}}     
\\  
\bottomrule
\end{tabular}
}
\label{tab:ablation_study}
\end{table}

\mypara{Transferability under label-efficient fine-tuning.}
\label{label_efficient_transfer_results}
For label-efficient transferabiluty evaluation, \nus{} training split is sampled with different ratios (1\% to 10\%) to generate label-efficient fine-tuning datasets. For fair comparisons, all models are end-to-end fine-tuned for the same iterations as the default fine-tuning setting (100\%).
 As shown in Table~\ref{tab:nuscene_label_efficient_det}, NS-MAE shows the best performance under all settings for 3D object detection and BEV-map segmentation.

\subsection{Component Effectiveness Study}
\label{ablation_study}

We conducted the ablation study using BEVFusion (image size $128\times352$, single-sweep LiDAR) as the baseline for 3D object detection. 
Table~\ref{tab:ablation_study} demonstrates how different components of NS-MAE contribute to overall performance. Notably, the interaction between multi-modal masking, viewpoint selection, and rendering enhances representation learning. By enforcing reconstruction through masking, NS-MAE avoids trivial solutions, while the careful selection of BEV and perspective viewpoints allows the network to learn complementary geometric and semantic features. 

\noindent\textbf{The importance of input modality masking.} 
Masking input data plays a crucial role in preventing the network from learning trivial identity mappings, which can occur without masking (A1, B3), resulting in performance degradation below random initialization. As demonstrated by the improvements in settings A2, B1, and B2, masking enhances the network’s ability to learn meaningful representations by enforcing the reconstruction of masked regions.

\noindent\textbf{Rendering with key viewpoints boosts pre-training performance.} The selection of viewpoint directions is vital for effective pre-training. Depth rendering from two critical viewpoints—perspective (B1) and bird's-eye view (B2)—outperforms random viewpoint sampling (B0), as perspective views capture region-of-interest object details while BEV provides a holistic understanding of scene geometry. This complementary information is essential for driving tasks, as it offers both spatial layout and detailed object-level semantics. Joint rendering from multiple views (B4) further enhances performance, with the default configuration (including rendering targets $\mathbf{C}$, $\mathbf{D}^{PER}$, and $\mathbf{D}^{BEV}$) achieving a 1.5\% higher NDS and 1.0\% higher mAP than baseline. This highlights the importance of carefully selecting and combining viewpoints during pre-training to maximize representation learning and generalization across tasks.

\section{Conclusion}
\label{sec:Conclusion}

In this work, we presented NS-MAE, a unified and scalable self-supervised pre-training strategy that effectively learns transferable multi-modal representations for 3D perception in autonomous driving. By combining masked reconstruction with NeRF-like rendering, NS-MAE unifies the optimization process across LiDAR and camera data, addressing limitations in modality-specific pre-training approaches. Our extensive experimental results demonstrate that NS-MAE achieves superior performance on various tasks, particularly in label-efficient fine-tuning settings. 

In \textbf{future work}, we aim to  to explore the scalability of this method on larger datasets such as Waymo~\cite{waymo2020} and Argoverse~\cite{wilson2023argoverse}, and investigate its application to additional sensor modalities like radar~\cite{yang2020radarnet}, further enhancing its versatility in real-world autonomous driving systems.


\bibliographystyle{IEEEtran}
\bibliography{main.bib}

\end{document}